\documentclass{article}

    \usepackage[preprint]{neurips_2023}

\usepackage{tikz}
\usepackage{graphicx}
\usepackage{adjustbox}
\usepackage{comment}
\usepackage{amsmath}
\usepackage{caption}
\usepackage{subcaption}
\usepackage{booktabs}

\usepackage[utf8]{inputenc} 
\usepackage[T1]{fontenc}    
\usepackage{hyperref}       
\usepackage{url}            
\usepackage{booktabs}       
\usepackage{amsfonts}       
\usepackage{nicefrac}       
\usepackage{microtype}      
\usepackage{xcolor}         

\title{Self-Supervised Learning from Non-Object Centric Images with a Geometric Transformation Sensitive Architecture}

\author{%
  Taeho Kim \quad Jong-Min Lee$^{*}$\\
  Department of Artificial Intelligence\\
  Hanyang University\\
  222, Wangsimni-ro, Seongdong-gu, Seoul, Republic of Korea \\
  \texttt{bok3948@hanyang.ac.kr \quad ljm@hanyang.ac.kr} \\
}

\begin{document}

\maketitle

\begin{abstract}
  Most invariance-based self-supervised methods rely on single object-centric images (e.g., ImageNet images) for pretraining, learning features that invariant to geometric transformation. However, when images are not object-centric, the semantics of the image can be significantly altered due to cropping. Furthermore, as the model becomes insensitive to geometric transformations, it may struggle to capture location information. For this reason, we propose a Geometric Transformation Sensitive Architecture designed to be sensitive to geometric transformations, specifically focusing on four-fold rotation, random crop, and multi-crop. Our method encourages the student to be sensitive by predicting rotation and using targets that vary with those transformations through pooling and rotating the teacher feature map. Additionally, we use patch correspondence loss to encourage correspondence between patches with similar features. This approach allows us to capture long-term dependencies in a more appropriate way than capturing long-term dependencies by encouraging local-to-global correspondence, which occurs when learning to be insensitive to multi-crop. Our approach demonstrates improved performance when using non-object-centric images as pretraining data compared to other methods that train the model to be insensitive to geometric transformation. We surpass DINO[\citet{caron2021emerging}] baseline in tasks including image classification, semantic segmentation, detection, and instance segmentation with improvements of 4.9 $Top-1 Acc$, 3.3 $mIoU$, 3.4 $AP^b$, and 2.7 $AP^m$. Code and pretrained models are publicly available at: \url{https://github.com/bok3948/GTSA}
\end{abstract}

\section{Introduction}
Invariance-based methods are one of the primary self-supervised learning approaches for computer vision. These methods learn to be insensitive to various transformations, such as rotations, flips, crops, color jittering, blurring, and random grayscale, which provide an inductive bias that helps with representation learning[\citet{Chen2020ExploringSS}, \citet{bardes2022vicreg}, \citet{Zbontar2021BarlowTS}].

Augmentations employed in self-supervised learning methods can be divided into two categories: photometric transformations and geometric transformations. Photometric transformations, such as color jittering, Gaussian blurring, and grayscale conversion, involve changes to the appearance of an image, like color, brightness, or contrast. geometric transformations, including random crop, multi-crop, flip and rotation, deal with changes to the spatial configuration of the image. 

In the case of pretraining with non-object centric image, Learning invariant features from crop-related geometric transformations can be problematic. This is because cropped views may not always depict the same object[\citet{purushwalkam2020demystifying}, \citet{zhang2022leverage}]. In contrast, object-centric images are less prone to such issues due to their inherent focus on specific objects, which remain semantically consistent across various augmentations. This explains the significant performance drop observed when applying invariant methods to non-object centric images[\citet{purushwalkam2020demystifying}, \citet{elnouby2021largescale}]. It can also be one of the reasons that, to obtain comparable results with curated datasets, a considerably larger amount of uncurated data is required[\citet{goyal2021selfsupervised}].

Furthermore, when learning to be insensitive to geometric transformations, there is a risk of not capturing location information, and dense prediction models need to be sensitive to these transformations rather than insensitive. Therefore, being insensitive 
 to geometric transformations may not be appropriate.

As mentioned earlier, training a model to be insensitive to geometric transformations may lead to noise in learning. However, these transformations can still be beneficial for representation learning since they prevent pathological training behavior[\citet{chen2020simple}] and provide diversity in inputs. Therefore, we propose a method that focuses on training a model to be sensitive with respect to those transformations, instead of insensitive to those transformations.

To achieve this, we must provide a target that varies according to the input's geometric transformation during training. to create a target that varies with cropping, we pool the overlapping region from the teacher's feature map, which can be seen as cropping, and provide it to the student as a target. Additionally, to make the model sensitive to four-fold rotations, we rotate the target feature map to align it appropriately with the student input, and we include a prediction task to predict the degree of rotation of the student's input.

Furthermore, we use a patch correspondence loss in our approach. When learning invariant features through multi-crop inputs, it will encourage global-to-local correspondence[\citet{caron2021emerging}, \citet{caron2021unsupervised}], resulting in the capture of long-term dependencies. Our model uses an additional loss that encourages correspondence between patch representations through cosine similarity, allowing us to capture long-term dependencies [\citet{bardes2022vicregl}]. Unlike encouraging correspondence between randomly selected crops, our approach induces correspondence between those that are similar in feature space, leading to more accurate correspondence.

Our experiments demonstrate that when using non-centric images as pretraining data, it is more advantageous to train a model to be sensitive to geometric transformations rather than insensitive. We significantly outperformed prominent invariance-based methods in various tasks, including image classification, semantic segmentation, detection, and instance segmentation.

\section{Related Work}
\label{gen_inst}
\textbf{Non-Contrative Learning}
Non-contrastive learning methods aim to learn an invariant bias towards transformations by training on different views of the same image, without explicit negative samples[\citet{garrido2022duality}]. In the absence of negative samples, non-contrastive learning methods employ various alternative approaches to prevent representation collapse. These include non-contrastive losses that minimize redundancy across embeddings[\citet{bardes2022vicreg},\citet{Zbontar2021BarlowTS}], clustering-based techniques that maximize the entropy of the average embedding[\citet{caron2019deep}, \citet{caron2021unsupervised}, \citet{goyal2021selfsupervised}, \citet{assran2022masked}], centering and sharpening output features[\citet{caron2021emerging}], and heuristic strategies utilizing asymmetric architectural design with stop-gradient, additional predictors, and momentum encoders[\citet{Richemond2020BYOLWE}, \citet{Chen2020ExploringSS}, \citet{tian2021understanding}]. Our method belongs to the non-contrastive learning category and adopts an asymmetric architectural design to prevent representation collapse.

\textbf{Self-Supervised Learning with Uncurated Dataset}
Several self-supervised pretraining methods have been proposed for uncurated datasets, such as the clustering-based method presented in [\citet{caron2021unsupervised}, \citet{tian2021divide}]. These methods have shown good performance even when using uncurated datasets, and [\citet{goyal2021selfsupervised}] demonstrated the scalability of their method to larger datasets for increased performance. Additionally, [\citet{elnouby2021largescale}] showed that, given sufficient iterations, even a small non-object centric dataset can yield results that are comparable to those obtained using a larger, highly curated dataset.

However, our approach differs from other methods that aim to adapt to uncurated datasets. While clustering-based techniques are used in these methods, they still learn invariant representations to augmentations, whereas our approach learns features that are sensitive to geometric transformations. On the other hand, [{\citet{tian2021divide}}] aims to address the shift in the distribution of image classes rather than object-centric bias.

\textbf{Self-Supervised Methods that Learn to be Sensitive to Transformations}
Early self-supervised learning methods, such as [\citet{noroozi2017unsupervised}, \citet{yamaguchi2021image}], train the model to be sensitive to transformations by predicting the permutation or rotation applied to the input. As contrastive learning has gained prominence in representation learning, the importance of learning transformation-invariant representations has become increasingly evident[\citet{misra2019selfsupervised}, \citet{chen2020simple}, \citet{he2020momentum}]. More recent work has utilized a hybrid approach that is sensitive to some transformations and insensitive to others [\citet{dangovski2022equivariant}]. Performance improvement has been achieved by training to be sensitive to four-fold rotation while insensitive to other transformations. Similarly, our model also learns to be sensitive to four-fold rotations and insensitive to other photometric transformations. However, our method additionally becomes sensitive to crop-related transformations.

\begin{figure}[t]
\centering
\adjustimage{bgcolor=white,width=0.8\textwidth}{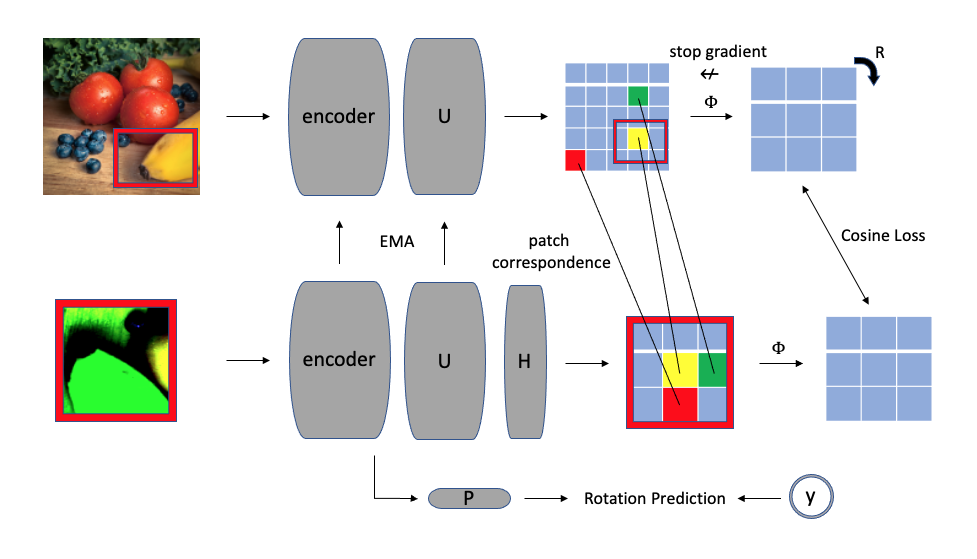}
\caption{\textbf{The GTSA.} The Geometric Transformation Sensitive Architecture (GTSA). the teacher only receives global views only, while the student receives both global and local views. The learning process is designed to increase the similarity in overlapping regions and predicting four-fold rotation. Additionally, to capture long-term dependencies, GTSA encourages similarity between the teacher's patch representations and the student's patch representations by matching these patch representations using cosine similarity.}
\label{fig:gtsa}
\end{figure}

\section{Methods}
\label{headings}
In this section, we describe the training procedure for our proposed GTSA method, as illustrated in Figure \ref{fig:gtsa}. We adopt an asymmetric Teacher-Student architecture, similar to those in [\citet{Richemond2020BYOLWE}, \citet{Chen2020ExploringSS}]. The Teacher comprises an encoder and a projector, while the Student includes an encoder, projector, and predictor. Following the multi-crop strategy used in [\citet{caron2021unsupervised}, \citet{caron2021emerging}], we feed only the global view to the Teacher and both global and local views to the Student. Our objectives involve maximizing the similarity between overlapping region representations and similar patch representations and predicting rotation.

\textbf{Inputs.}
Similar to [\citet{ziegler2022selfsupervised}], we utilized various augmentation techniques including color jitter, random grayscale, random Gaussian noise, Gaussian blur, random resize crop, and multi-crop. Additionally, we employed four-fold rotation. For each image, we apply random augmentations and generate G global views and L local views. The inputs are $[x_1^{g}, x_{2}^{g}, \dots, x_{1}^{l}, x_{2}^{l}, \dots]$, where $g$ and $l$ indicate global and local view, respectively. $x$ represents batchified images, and $x \in \mathbb{R}^{B \times C \times H \times W}$. Here, $B$ is the batch size, $C$ represents the number of image channels, and $H$ and $W$ denote the image size.

\textbf{Teacher and Student.}
Apart from the additional predictor attached to the Student, the Teacher and Student share the same structural design. We utilized the ViT[\citet{dosovitskiy2021image}] as the encoder and employed stacked CNN blocks for the projector, each block comprising a convolution layer, layer normalization[\citet{ba2016layer}], GELU activation[\citet{hendrycks2020gaussian}] and residual connection[\citet{he2015deep}]. The predictor has a similar architecture to the projector but uses fewer CNN blocks in its composition. We annotated the Predictor and Projector as H and U, respectively. Note that both the projector and predictor do not reduce the spatial resolution of the encoder's output. The Teacher does not get updated through gradient descent; instead, its weights follow the Student's weights using exponential moving average.[\citet{tarvainen2018mean}, \citet{He2019MomentumCF}]

\textbf{Correspondence Region Pooling Operator.}
We introduce a correspondence region pooling operator, denoted as $\Phi(\cdot)$. To be sensitive to crop-related augmentations, the student must receive a target that reflects the crop augmentation. The $\Phi(\cdot)$ operator serves this purpose by cropping specific locations in the feature space. It extracts the overlapping portions between the teacher view and student view in the feature space.

To accomplish this, we first calculate the overlap region bounding boxes in the input space and scale them to match the feature spatial resolution. We then apply the $\Phi(\cdot)$ operator to both the student and teacher feature maps. We implement this operator using RoI-Align [\cite{he2018mask}].

\textbf{Rotation Operator.}
To be sensitive to rotation, we propose a rotation operator. This operator rotates the teacher output feature map according to the input rotation. We denote this as $R(\cdot)$. By using this operator, the student receives a target that reflects the input rotation.  

\textbf{Rotation Predictor.}
As we employ the rotation prediction pretext task, we extract a vector from the student encoder output using a Global Average Pooling layer[\citet{lin2014network}]. This vector is then input into the rotation predictor, which generates logits for the rotation prediction pretext task. The architecture of this process includes a linear layer, GELU activation, and a normalization layer. We denote the rotation predictor as P.

\textbf{Loss Function.}
We denote the output feature map of the student predictor as $z$ and the output feature map of the teacher's projection layer as $\tilde{z}$. $z \in \mathbb{R}^{B \times D \times h_s \times w_s}$ and $\tilde{z} \in \mathbb{R}^{B \times D \times h_t \times w_t}$, where $B$ is the batch size, $D$ is the feature dimension, $h_s$ and $w_s$ represent the spatial size of the student feature map, and $h_t$ and $w_t$ represent the spatial size of the teacher feature map. We apply $\Phi$ to both $z$ and $\tilde{z}$, and additionally apply $R$ to $\Phi(\tilde{z})$. Both $\Phi(z)$ and $R(\Phi(\tilde{z}))$ will have the same dimensions. $\Phi(z), R(\Phi(\tilde{z})) \in \mathbb{R}^{B \times D \times h_o \times w_o}$. We then compute the cosine similarity between them along feature dimension. The equation is as follows:

\begin{equation}
l(z, \tilde{z}) = -\frac{1}{B\times T} \sum_{i=1}^B \sum_{t=1}^T \frac{\Phi(z_{i})_{t} \cdot R(\Phi(\tilde{z}_{i}))_{t}}{\lVert \Phi(z_{i})_{t} \rVert \lVert R(\Phi(\tilde{z}_{i}))_{t} \rVert}
\end{equation}

where $i$ is the index of the batch, $t$ indicates the t-th spatial location in the output of the operator and $T$ is $(h_o\times w_o)$. the cosine similarity is calculated as the dot product of the two vectors divided by the product of their magnitudes.

For the Rotation prediction pretext task, we use rotation prediction loss, which is annotated with $l_{rp}$. We design this loss using cross-entropy loss. The equation is as follows:

\begin{equation}
l_{rp}(\mathbf{y}, \mathbf{\hat{y}}) = -\frac{1}{B} \sum_{i=1}^{B}  \mathbf{y}_{i} \cdot \log \mathbf{\hat{y}}_{i}
\end{equation}

where $\mathbf{y}_{i}$ is the target indicating the $i$-th sample rotation angle. $\mathbf{\hat{y}}_{i}$ is the Rotation Predictor output probability distribution of the $i$-th sample. In this case, as we use four-fold rotation, the possible rotation angles are [0, 90, 180, 270] degrees.

Additionally, we use patch correspondence loss, which we annotate with $l_{pc}$. We pair semantically similar patches from the teacher and student patch representations based on their cosine similarity and make them more alike using the $l_{pc}$ loss. However, as noise may exist in this process, we do not encourage correspondence between all patches. Instead, we only encourage the similarity of the top-$K$ most similar features among all patches, the same as  [\citet{bardes2022vicregl}]. This helps to ensure that we are aligning semantically similar patches while avoiding pairing patches that are too dissimilar.

\begin{equation}
l_{pc}(z, \tilde{z}) = -\frac{1}{B \times K} \sum_{i=1}^B \sum_{p=1}^K \frac{z_{i,p} \cdot \tilde{z}_{i,\tilde{p}}}{\lVert z_{i,p} \rVert \lVert \tilde{z}_{i, \tilde{p}} \rVert}
\end{equation}

Here, $z_{i,p}$ refers to the $p$-th patch representation of the i-th sample from the student, and $\tilde{z}_{i,\tilde{p}}$ denotes The patch representation that is the closest to $z_{i,p}$ among the patch representations for the teacher's i-th sample. $K$ represents  total number of matched pairs which are filtered.

To consider multi-crop scenarios, we use the following total loss function:

\begin{align}
\text{Loss} = &\frac{1}{G(L+G)-G}\sum_{g=1}^{G} \sum_{\substack{l=1\\ l\neq g}}^{L+G} l(z_{l}, \tilde{z}_{g}) + \alpha * \frac{1}{G(L+G)-G}\sum_{g=1}^{G} \sum_{\substack{l=1\\ l\neq g}}^{L+G} l_{pc}(z_{l}, \tilde{z}_{g}) \nonumber \\
&+ \beta * \frac{1}{L+G} \sum_{l=1}^{L+G} l_{rp}(\mathbf{y_{l}}, \mathbf{\hat{y_{l}}})
\end{align}

Here, $\alpha$ and $\beta$ are hyperparameters that control the impact of $l_{pc}$ and $l_{rp}$, respectively. $G$ and $L$ denote the total number of global views and local views, respectively.

\begin{table}[t]
\centering
\caption{\textbf{The iNaturalist 2019 image classification performance.} performance was obtained by pretraining all models with 100 epochs on COCO train2017, followed by fine-tuning for 300 epochs using the same settings for all models.}
\label{tab:classification}
\begin{tabular}{lcccc}
\toprule
Method  &Top-1 Acc & Top-5 Acc \\
\midrule

rand init   & 40.6 & 69.0 \\
MoCo v3   & 48.8 & 76.3 \\
DINO   & 54.8 & 82.9 \\

GTSA (Ours)    &\textbf{59.7} & \textbf{85.7} \\

\bottomrule
\end{tabular}
\end{table}

\section{Experiments}
\label{gen_inst}
Experiments have three distinct subsections for a comprehensive explanation of our approach.

In Section 4.1, we demonstrate the effectiveness of GTSA in learning high-quality representations from non-object centric images. We pretrained our model on the COCO train2017 dataset [\citet{lin2015microsoft}], a collection composed of non-object centric images. Then, we evaluated the performance of our model by fine-tuning it on various downstream tasks, specifically Classification, Detection, Instance Segmentation, and Semantic Segmentation. In addition, we pretrained our model on the ADE20K train dataset [\citet{zhou2018semantic}], which also consists of non-object centric images. However, due to the larger size of other downstream task datasets compared to the pretraining dataset, we restricted our evaluation to Semantic Segmentation on the ADE20K dataset.

In Section 4.2, we demonstrated that our model operates as intended, showcasing its sensitivity to rotation and crop-related transformations.

In Section 4.3, we compiled the results from our ablation study. We illustrated the effects of rotation prediction loss and patch correspondence, and through visualization of encouraged patch pairs, we verified that our method encourages correspondence even with patches that are far apart.

\textbf{Pretrain Setup.}
We used the same hyperparameters as DINO, as much as possible. Specifically, we set the batch size to 512, the global view size to 224x224, and the local view size to 96x96. We also used a scheduler to start the momentum at 0.996, just like DINO, and gradually increased it to 1. For the optimizer, we employed AdamW[\citet{loshchilov2019decoupled}] and set both $\alpha$ and $\beta$ to 0.5. When pretraining with the ADE20K train dataset, we changed the jitter strength, which is a hyperparameter used for color jittering, from 1.0 to 0.2 and didn't normalize the encoder's output, and also $\beta$ to 0.25. Apart from these differences, all other settings were the same. Our default model is ViT-S/16, and we pretrained it for 100 epochs with 8 NVIDIA GeForce RTX 3090 GPUs.

\begin{table}[t]
\caption{\textbf{The COCO 2017 detection and instance segmentation performance.} Performance was obtained by pretraining all models with 100 epochs on COCO train2017, followed by fine-tuning for a standard $1\times$ schedule using the same settings for all models.}

\label{tab:detection}
\centering
\begin{tabular}{lccccccc}
\toprule
Method  &\multicolumn{3}{c}\centering {\hspace{3em}Detection} &  &\multicolumn{3}{c}{Instance Segmentation} \\
\cmidrule(lr){3-5} \cmidrule(lr){6-8}
& & $AP\textsuperscript{b}$ & $AP\textsuperscript{b}_{50}$ & $AP\textsuperscript{b}_{75}$ & $AP\textsuperscript{m}$ & $AP\textsuperscript{m}_{50}$ & $AP\textsuperscript{m}_{75}$ \\
\midrule
rand init & &23.2 &42.3 &22.5 &23.0 &40.0&23.3\\
MoCo v3 &  & 29.6 & 50.1 & 30.4 & 28.3 & 47.7 &29.2 \\
DINO &  & 32.4 & 54.2 & 33.8 & 30.8 & 51.1 &32.2 \\
GTSA(ours) &  & \textbf{35.8} & \textbf{57.8} & \textbf{38.5} & \textbf{33.5} & \textbf{54.7} & \textbf{35.3} \\
\bottomrule
\end{tabular}
\end{table}

\begin{table}[t]
\centering
\caption{\textbf{The ADE20K image semantic segmentation performance.} Performance was obtained by pretraining all models with 100 epochs on COCO train2017, followed by fine-tuning for a standard 40K iteration schedule.}
\label{tab:semantic}
\begin{tabular}{lccc}
\toprule
Method & aAcc & mIoU & mAcc \\
\midrule
rand init & 64.5 & 12.1 & 16.1 \\
MoCo v3 & 72.7 & 23.5 & 31.7 \\
DINO & 74.7 & 27.3 & 35.9 \\
GTSA (Ours) & \textbf{76.4} & \textbf{30.6} & \textbf{40.0} \\
\bottomrule
\end{tabular}
\end{table}

\subsection{Fine-tuning}

\textbf{Baseline.}
We chose Dino and MoCo v3[\citet{chen2021empirical}] as our baseline methods. These methods are highly relevant to our research as they, like us, employ the Vision Transformer (ViT) as an encoder and focus on training the model to be insensitive in a self-supervised manner. Thus, they provide a compelling counterpoint to our approach which train the model to be sensitive to geometric transformations.

The official codes from these baseline methods were leveraged in producing our results. To ensure a fair and balanced comparison, all methods underwent pretraining under the same conditions: 100 pretraining epochs and the use of the ViT-S/16 encoder. Moreover, the fine-tuning process was executed in an entirely same manner across all methods.

\textbf{Image Classification.}
We compared our method with other self-supervised methods in terms of image classification performance when fine-tuning on the iNaturalist 2019 dataset [\citet{vanhorn2018inaturalist}]. Table \ref{tab:classification} shows that our method outperforms other methods that learn only invariant features. We achieve a 4.9 and 10.9 higher accuracy compared to DINO and MoCo-v3, respectively, and a 19.1 accuracy improvement compared to random initialization.

\textbf{Detection and Instance Segmentation.}
Table \ref{tab:detection} shows the performance of our method on COCO detection and instance segmentation tasks. GTSA outperforms DINO and MoCo-v3 by 3.4 and 6.2 $AP^{b}$ in detection, and by 2.7 and 5.2 $AP^{m}$ in instance segmentation, respectively. All models were fine-tuned using Mask R-CNN [\citet{he2018mask}] and FPN [\citet{lin2017feature}] under the standard 1x schedule.

\begin{table}[t]
\centering
\caption{\textbf{The performance of ADE20K image semantic segmentation when pretraining on the ADE20K train dataset.} Performance was obtained by pretraining all models with 100 epochs on ADE20K train dataset, followed by fine-tuning for a standard 40K iteration schedule.}
\label{tab:16}
\begin{tabular}{lccc}
\toprule
Method & aAcc & mIoU & mAcc \\
\midrule
rand init & 64.5 & 12.1 & 16.1 \\
MoCo v3 & 66.3 & 13.6 & 19.0 \\
DINO & 65.5 & 13.6 & 18.8 \\
GTSA (Ours) & \textbf{68.4} & \textbf{16.2} & \textbf{22.6} \\
\bottomrule
\end{tabular}
\end{table}

\begin{figure}[t]
\centering
\adjustimage{bgcolor=white,width=0.8\textwidth}{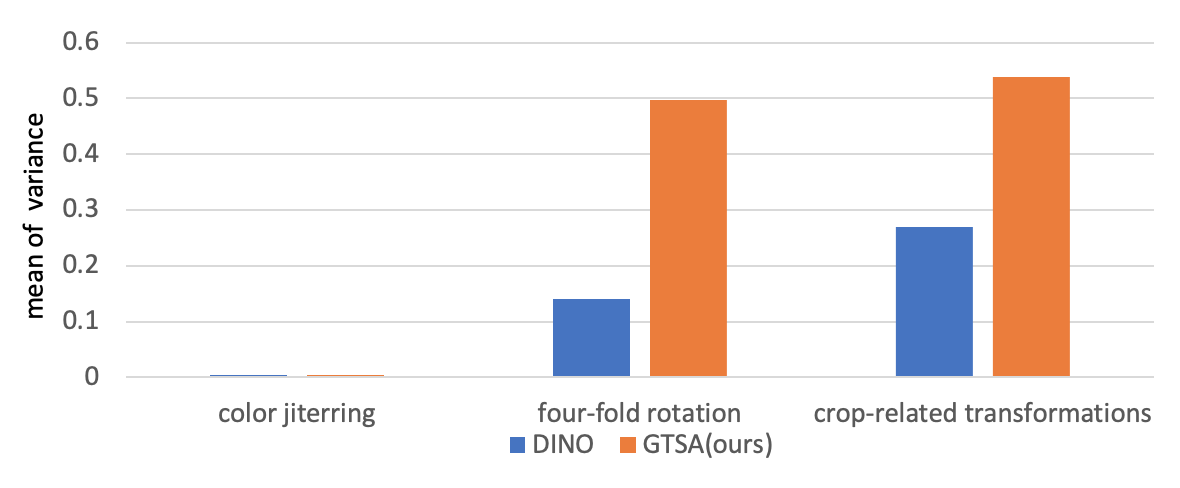}
\caption{\textbf{Output variance comparison.} This figure demonstrates how sensitive the DINO and our model are to input transformations. The y-axis represents the mean of output variance of the pretrained encoder. It demonstrates that our model is more sensitive to geometric augmentations than DINO.}
\label{fig:variance}
\end{figure}

\begin{table}[t]
\centering
\caption{\textbf{Ablation Study: effect of $\mathbf{l_{pc}}$ and $\mathbf{l_{rc}}$} This table shows that the segmentation performance improves with the additional use of ${l_{pc}}$ and ${l_{rc}}$.}
\label{tab:effect}
\begin{tabular}{lcccc}
\toprule
loss function & aAcc & mIoU & mAcc \\
\midrule
$l$ & 67.3 & 15.4 &21.5 \\
$l$ +  $l_{pc}$ &67.7 & 15.8 & 21.7  \\
$l$ +  $l_{pc}$ + $l_{rc}$ & \textbf{68.4} & \textbf{16.2} & \textbf{22.6} \\
\bottomrule
\end{tabular}
\end{table}

\begin{figure}[t]
\centering
\adjustimage{bgcolor=white,width=0.8\textwidth}{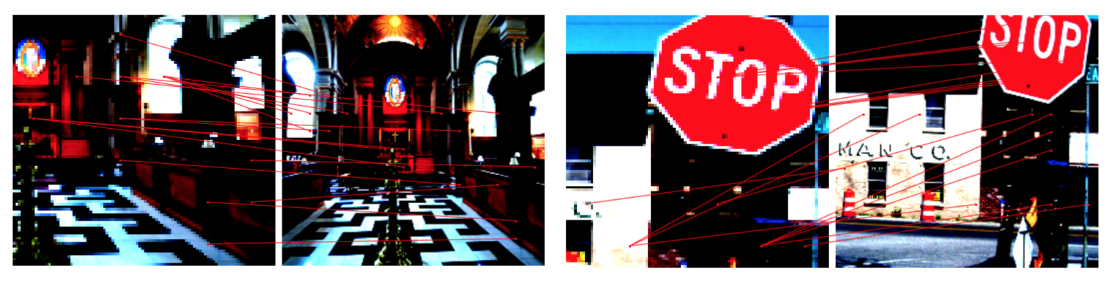}
\caption{\textbf{Ablation Study: Matched Pairs.} The visualization of matched pairs when computing patch $l_{pc}$. The visualization was generated by feeding images from the COCO val2017 dataset into a GTSA model, which had been previously trained on the COCO train2017 dataset.}
\label{fig:matched pairs}
\end{figure}

\textbf{Semantic Segmentation.}
Table \ref{tab:semantic} reports the performance of on ADE20K semantic segmentation using the Acc, mIoU, and mAcc metrics with all methods are pretrained with COCO train2017 dataset. While DINO achieves a 27.3 mIoU, GTSA attains a higher performance of 30.6 mIoU, which is a 3.3 mIoU improvement. Moreover, our method outperforms MoCo-v3 by 7.1 mIoU. All models were fine-tuned using Semantic FPN [\citet{kirillov2019panoptic}] under the standard 40k iteration schedule, following the same approach as in [\citet{yun2022patchlevel}].

Table \ref{tab:16}, also reports the performance of on ADE20K semantic segmentation but differ in that in this table use ADE20K train dataset as pretraining data. GTSA outperforms DINO and MoCo with inprovement 2.6 mIoU. fine-tuning setting are all same to Table \ref{tab:semantic}.

\subsection{Proving sensitivity to geometric transformations}
\label{4.2}
In this section, we present Figure \ref{fig:variance} to showcase our model's sensitivity to various transformations. We designed this experiment by measuring the variance in the output based on input transformations. Specifically, we generated ten views with single type of augmentation and fed these views into a model pretrained using the DINO method and another pretrained using our method. We then measured the variance of the encoder output generated by global average pooling. Here, we utilized the COCO val2017 dataset to compute the mean of variance, which is denoted on the y-axis.

As shown in Figure \ref{fig:variance}, both DINO and GTSA learned to be invariant to color jittering, resulting in a very low variance. However, for four-fold rotation and crop-related transformations, GTSA exhibited a substantially higher variance compared to DINO. The crop-related transformations was implemented using a random resize crop, creating two global views and eight local views. It is important to note that the exact same inputs were fed into both the DINO and GTSA.

\subsection{Ablation study.}
In this section, we demonstrate the performance enhancement achieved through $l_{pc}$ and $l_{rp}$, and we present a figure that visualizes matched pairs. This illustrates that even distant patches are matched, confirming that correspondence is encouraged over long distances.

As displayed in Table \ref{tab:effect}, we set $l$ as the baseline and showed performance improvement with the addition of $l_{pc}$ and $l_{rp}$. For simplicity, we pretrain on the ADE20K train dataset for 100 epochs and report the results for Semantic Segmentation on the ADE20K dataset. We observed a 0.4 mIoU increase upon adding $l_{pc}$, and an additional 0.4 mIoU increase when $l_{rp}$ was incorporated.

Figure \ref{fig:matched pairs} visualizes matched pairs. We used GTSA, which was pretrained for 100 epochs on the COCO train2017 dataset, and input images from COCO val2017 to obtain the matched pairs. From the left image, we can see that matching occurs between parts that depict columns, even if they are not precisely the same column. Similarly, in the right image, we see matching between two parts, both depicting a wall, despite being located at a distance from each other. This demonstrates that our method encourages the capture of long-term dependencies as we intended.

\section{Conclusion}
We propose the Geometric Transformation Sensitive Architecture (GTSA) as a self-supervised method designed for non-object centric images.
Our approach trains the model to be sensitive to geometric transformations, specifically rotation and crop-related transformations, by utilizing targets that reflect geometric transformations. Experimental results demonstrate that our method outperforms other transformation-invariant methods when pretrained on non-object centric images.

\textbf{Limitations and Future Works.} Our method does not learn to be sensitive to all types of geometric transformations. Specifically, it is trained to be sensitive to four-fold rotations and crop-related transformations. In the future, we aim to explore its effectiveness when made sensitive to a broader range of geometric transformations. Moreover, we will conduct research to achieve superior performance on curated datasets.

\bibliographystyle{plainnat}
\bibliography{reference}

\end{document}